\documentclass{ceurart}

\usepackage{acro}
\usepackage{tikz}
\usepackage{flafter}
\usepackage[font=small,labelfont=bf,textfont=normalfont]{caption}
\usepackage[font=small,labelfont=bf]{subcaption}
\usetikzlibrary{positioning,arrows.meta}
\DeclareAcronym{kg}{short=KG,long=knowledge graph}
\DeclareAcronym{ad}{short=AD,long=autonomous driving}
\DeclareAcronym{owl}{short=OWL,long=Web Ontology Language}
\DeclareAcronym{ttc}{short=TTC,long=time to collision}
\DeclareAcronym{ttl}{short=TTL,long=Terse RDF Triple Language}
\DeclareAcronym{sparql}{short=SPARQL,long=SPARQL Protocol and RDF Query Language}
\DeclareAcronym{kge}{short=KGE,long=knowledge graph embedding}
\DeclareAcronym{tbox}{short=T-Box,long=terminological box}
\DeclareAcronym{abox}{short=A-Box,long=assertional box}

\copyrightyear{2026}
\copyrightclause{Copyright for this paper by its authors. Use permitted under Creative Commons License Attribution 4.0 International (CC BY 4.0).}
\conference{Joint Workshop on Statistics and Knowledge Integration for Logic, Learning, Ethical Decisions, and LLMs (SKILLED-LLMs 2026), co-located with FLoC 2026, Lisbon, Portugal}

\title{A Semantic Knowledge Graph Construction Pipeline for Vehicle Intention Prediction in nuScenes}
\author[1]{Khaled Nowara}[%
email=k.nowara02@gmail.com,
]

\author[2]{Mohamed Manzour}[%
email=ahmed.manzour@uah.es,
]

\author[1]{Catherine Elias}[%
email=catherine.elias@guc.edu.eg,
]

\address[1]{German University in Cairo, Cairo, Egypt}
\address[2]{Computer Engineering Department, University of Alcalá, Madrid, Spain}

\begin{document}

\begin{abstract}
Knowledge graphs (KGs) are increasingly used in autonomous driving (AD) to make traffic
scene entities and relations explicit for prediction and reasoning. Existing
traffic-scene KGs, however, primarily emphasize trajectory forecasting or
general scene understanding. This paper presents the \emph{nuScenes Intention Knowledge
Graph} (NIKG), a focused Web Ontology Language (OWL)-based semantic representation and construction
pipeline for ego-centric vehicle intention context. NIKG instantiates nuScenes
scenes as a self-describing Turtle graph and promotes three intention-relevant
relations to first-class properties of each scene participant: ego-relative
motion, spatial configuration, and time to collision (TTC) risk. The pipeline uses
a small input contract available in many ego-centric driving datasets: ego
pose, tracked object annotations, and temporal keyframe order.
NIKG is released as an ontology-backed, self-describing graph artifact together with a reproducible construction pipeline and validation protocol for ego-centric intention context.
Applying this pipeline to nuScenes mini yields a graph of 211{,}335 triples across 404 keyframes; under the validation protocol, ego-relative motion labels achieve 92.5\% forward consistency.
Coverage, semantic consistency, and temporal coherence analyses over this artifact are further reported.
NIKG occupies the space between scene-oriented traffic
KGs and intention-prediction models: a small, explicit semantic layer
intended to support downstream work such as knowledge graph embedding (KGE),
SPARQL-based feature extraction, or hybrid neuro-symbolic reasoning.
\end{abstract}


\maketitle

\section{Introduction}

Anticipating the behavior of surrounding traffic agents is one of the most challenging aspects of autonomous driving.
Human drivers do this constantly and largely unconsciously: they combine perception with contextual reasoning to judge whether a nearby vehicle is preparing to change lanes, likely to yield at a merge, or about to brake.
Replicating this capability---predicting agent \emph{intentions} rather than merely their immediate positions---is essential for safe planning in complex, interactive traffic
\cite{weiIntentionbasedRiskAwareTrajectory2025,manzourVehicleLaneChange2023}.

Data-driven approaches have made substantial progress on trajectory prediction, yet producing reliable intention estimates requires more than raw motion history.
It depends on richer contextual reasoning over the scene---the kind of semantic understanding that goes beyond raw trajectories and captures the relational and situational structure of the traffic environment.
Such context is typically computed implicitly inside learned encoders, making it hard to inspect, audit, or reuse across different models and studies \cite{zhouPedestrianIntentionEstimation2024}.

\Acp{kg} offer a principled way to make scene context explicit and structured.
By encoding entities, relationships, and semantic constraints in an ontology-grounded format, \acp{kg} provide a query-friendly representation that complements learned perception with symbolic reasoning and knowledge reuse
\cite{wickramarachchiKnowledgeGraphsDriving2024,wickramarachchiEvaluationKnowledgeGraph2020}.
Applied to autonomous driving, \acp{kg} have been used to represent scene topology, power \ac{kge} experiments over agent interaction data, and embed scene semantics for trajectory prediction \cite{mlodzianNuScenesKnowledgeGraph2023a,sunSemanticFormerHolisticSemantic2024}.

Despite this progress, no existing resource makes the contextual signals most relevant to intention prediction---ego-relative motion, spatial configuration, and \ac{ttc} risk---available as explicit, ontology-grounded facts tied to dataset keyframes.
These signals remain internal to learned models rather than published as reusable semantic content, forcing every new study to re-derive the same geometric inferences from raw annotations.

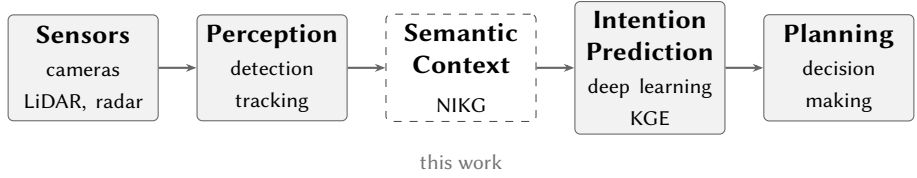
\begin{figure}[t]
\centering
\begin{tikzpicture}[
    node distance=3mm and 5mm,
    box/.style={
        rectangle, rounded corners=2pt, draw=black!60, fill=gray!10,
        text width=1.75cm, minimum height=1.4cm,
        align=center, font=\small
    },
    nikg/.style={
        rectangle, rounded corners=2pt, draw=black!70, dashed, fill=white,
        text width=1.75cm, minimum height=1.4cm,
        align=center, font=\small
    },
    arrow/.style={-{Stealth[length=4pt]}, thick, draw=black!60}
]
\node[box] (sensors)    {\textbf{Sensors}\\[3pt]\scriptsize cameras\\LiDAR, radar};
\node[box,  right=of sensors]    (perception) {\textbf{Perception}\\[3pt]\scriptsize detection\\tracking};
\node[nikg, right=of perception] (nikg)       {\textbf{Semantic\\Context}\\[3pt]\scriptsize NIKG};
\node[box,  right=of nikg]       (prediction) {\textbf{Intention\\Prediction}\\[3pt]\scriptsize deep learning\\KGE};
\node[box,  right=of prediction] (planning)   {\textbf{Planning}\\[3pt]\scriptsize decision\\making};

\draw[arrow] (sensors)    -- (perception);
\draw[arrow] (perception) -- (nikg);
\draw[arrow] (nikg)       -- (prediction);
\draw[arrow] (prediction) -- (planning);

\node[font=\scriptsize, text=black!55, below=3mm of nikg] {this work};
\end{tikzpicture}
\caption{NIKG in the autonomous driving pipeline. The semantic context layer
(dashed, this work) sits between perception and the downstream intention
predictor, making ego-relative motion, spatial configuration, and \ac{ttc}
risk available as explicit, queryable graph facts.}
\label{fig:pipeline-position}
\end{figure}

This paper introduces the \emph{nuScenes Intention Knowledge Graph} (NIKG) to fill that gap.
NIKG promotes those three signals to first-class, ontology-grounded properties on each observed agent, building on the structural scene representation of prior nuScenes graph work \cite{mlodzianNuScenesKnowledgeGraph2023a}.
Figure~\ref{fig:pipeline-position} shows where NIKG sits in the autonomous driving processing stack: between perception and the downstream intention predictor, as an explicit semantic context layer.
Section~2 situates NIKG relative to prior work and states the contributions; Sections~3--4 present the methodology and validation; Section~5 concludes.

\section{Related Work}

NIKG is situated within two areas: \acp{kg} as a semantic modeling layer for
\ac{ad} scenes (\S\ref{subsec:rw-kg-ad}), and the contextual cues used in maneuver-intention prediction (\S\ref{subsec:rw-intention}).
The resulting gaps are then summarized and NIKG's positioning is stated explicitly (\S\ref{subsec:rw-gap}).

\subsection{Knowledge graphs as a semantic layer in autonomous driving}
\label{subsec:rw-kg-ad}

Human drivers predict behavior by combining external perceptual cues with
internalized knowledge about context and intent. Data-driven approaches excel
at the perceptual side but struggle to represent the implicit semantic
structure that human drivers use \cite{zhouPedestrianIntentionEstimation2024}.
\Acp{kg} address this gap directly: they provide a structured mechanism for
representing, organizing, reasoning over, and retrieving semantic
relationships that would otherwise remain implicit in raw data
\cite{yangReviewKnowledgeGraph2018}.

This modeling pattern has been applied across several subproblems in
\ac{ad}. \Acp{kg} have been used to represent urban traffic topology and
support knowledge reuse via reasoning \cite{tanKnowledgeGraphUrban2021}, to
manage the heterogeneous sensor data generated by autonomous vehicles
\cite{wickramarachchiEvaluationKnowledgeGraph2020}, and to evaluate
\acp{kge} for \ac{ad} scene data
\cite{wickramarachchiKnowledgeGraphsDriving2024}. Together these demonstrate
that \acp{kg} are a broadly applicable semantic layer across \ac{ad}
subproblems, not only a tool for a single task.

For traffic scene semantics and prediction specifically, nSKG provides the
closest structural precedent: a semantic representation of nuScenes scenes
used for trajectory prediction \cite{mlodzianNuScenesKnowledgeGraph2023a}.
SemanticFormer embeds graph-structured scene semantics directly inside a
trajectory-prediction model \cite{sunSemanticFormerHolisticSemantic2024},
while FM4SU serializes a \ac{kg} into token sequences for language-model-style
scene understanding \cite{zhouPredictingRoadAhead2025a}. Pedestrian-oriented
work further shows that explicit graphs can support behavior interpretation
and not only trajectory forecasting
\cite{zhouPedestrianIntentionEstimation2024,meloKnowledgebasedExplainablePedestrian2024}.
However, these efforts target trajectories, pedestrian behavior, or embed the
graph inside a specific model architecture. None release a standalone,
vehicle-intention-oriented resource with an explicit ontology aligned to
maneuver context.

\subsection{Maneuver intention context}
\label{subsec:rw-intention}

The maneuver-intention prediction literature consistently decomposes context
into ego-relative surround layout, longitudinal and lateral positioning, and
scene-level risk. The surround-grid representation, in which the space around
the ego vehicle is divided into an ego-relative occupancy layout, is a widely
adopted encoding for this purpose
\cite{deoConvolutionalSocialPooling2018,manzourVehicleLaneChange2023,weiIntentionbasedRiskAwareTrajectory2025}.
\Ac{ttc} is a classical safety metric that quantifies collision risk from
relative closing speed and distance, and appears alongside positional layout
as a core contextual input in maneuver-intention pipelines
\cite{haywardTimeCollisionTechnique1972,weiIntentionbasedRiskAwareTrajectory2025}.

Beyond \ac{ad}, cross-domain work in service robotics shows that combining
explicit knowledge structures with intention reasoning is a reusable modeling
pattern \cite{zhouEnhancingIntentionPrediction2024}, further motivating a
\ac{kg}-based formalization of these factors.

\subsection{Gap summary and NIKG positioning}
\label{subsec:rw-gap}

Taken together, prior work suggests two complementary gaps.
First, semantic graphs built from driving datasets are often tuned toward trajectory prediction or fused into a specific neural architecture, rather than offered as a compact, standalone resource that downstream groups can adopt without re-deriving the graph \cite{mlodzianNuScenesKnowledgeGraph2023a,sunSemanticFormerHolisticSemantic2024,zhouPredictingRoadAhead2025a}.
Second, maneuver-intention pipelines routinely exploit ego-relative surround layout, motion relative to the ego, and \ac{ttc}-style risk \cite{deoConvolutionalSocialPooling2018,manzourVehicleLaneChange2023,weiIntentionbasedRiskAwareTrajectory2025}, yet these cues typically remain internal to learned encoders rather than published as explicit, ontology-grounded facts tied to dataset keyframes.

NIKG sits at this intersection.
NIKG reuses the nuScenes scene-graph scaffold from nSKG \cite{mlodzianNuScenesKnowledgeGraph2023a} and extends it with three interoperable relations---\texttt{motionTowardEgo}, \texttt{spatialConfiguration}, and \texttt{ttcRisk}---so that intention-relevant context becomes first-class, machine-checkable \ac{kg} content rather than an undocumented side effect of any one model.
The ontology, pipeline, and validation artifact are released so that downstream work---via \ac{kge}, \ac{sparql}-based features, or hybrid pipelines---can build directly on this representation without replicating the geometric preprocessing.

The paper makes three contributions:
\begin{enumerate}
\item a self-describing \ac{owl} ontology that elevates \texttt{motionTowardEgo}, \texttt{spatialConfiguration}, and \texttt{ttcRisk} to first-class relations on \texttt{SceneParticipant};
\item a modular construction pipeline from nuScenes to one merged \ac{ttl} graph encoding both schema and instances; and
\item an empirical characterization of the released artifact on nuScenes mini (coverage, semantic-consistency implications, and forward coherence of motion labels).
\end{enumerate}

\section{Methodology}

NIKG is described in two parts: the ontology that defines the semantic schema, and the pipeline that instantiates that schema on nuScenes and exports the resulting graph artifact.

\subsection{Ontology}

The ontology defines NIKG's semantic schema.
It is intentionally narrow: it captures ego-centric intention context rather than a complete model of traffic worlds.
The design goals are (i)~a self-describing schema with explicit classes and closed enumerations wherever relations take categorical values, (ii)~alignment with the structural vocabulary inherited from nSKG so that the scene graph scaffold remains comparable and interoperable \cite{mlodzianNuScenesKnowledgeGraph2023a}, and (iii)~release as a single reusable artifact that merges \ac{tbox} and \ac{abox} triples for downstream consumers.

\begin{figure*}[t]
\centering
\includegraphics[width=0.92\textwidth]{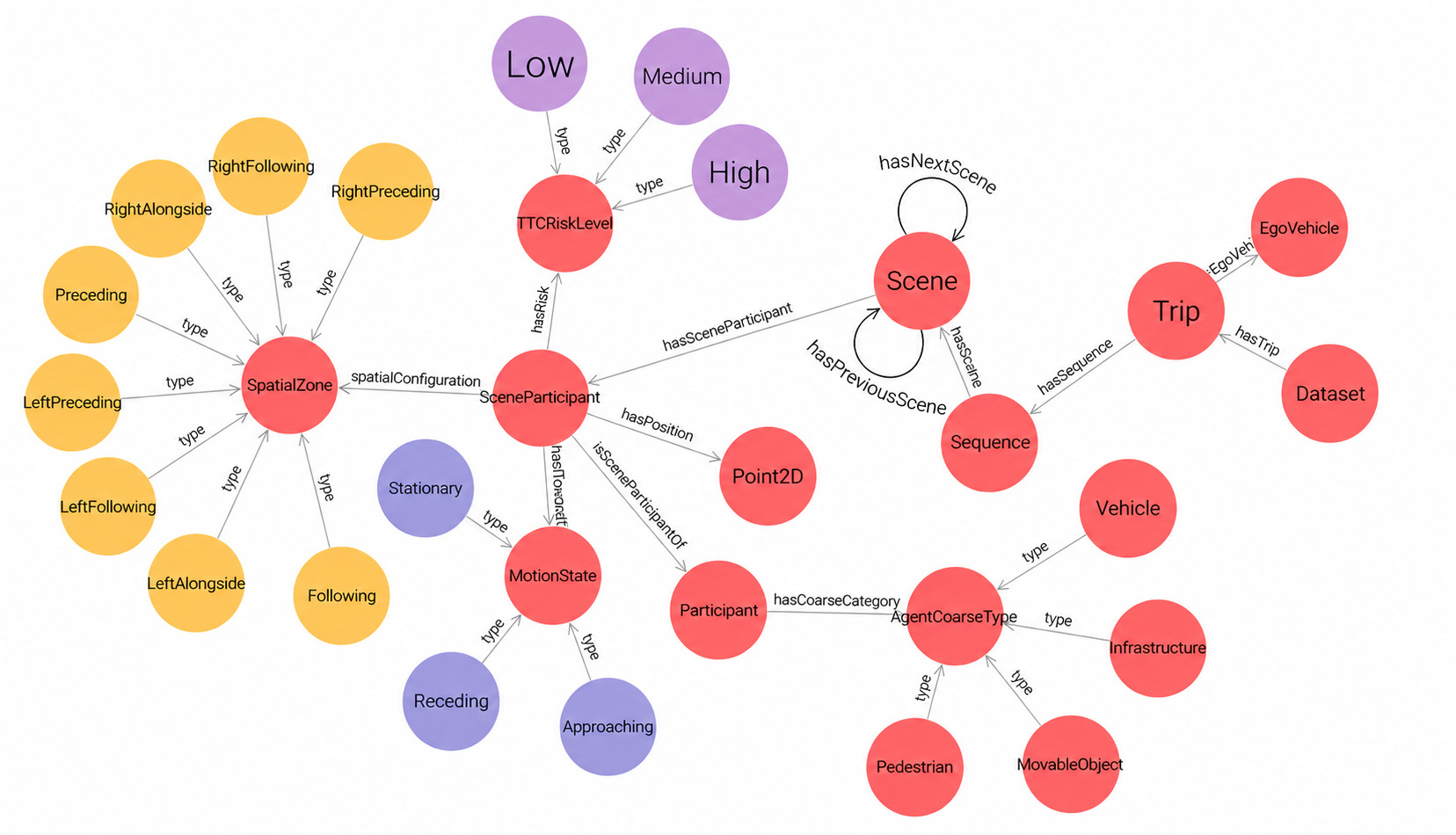}
\caption{NIKG ontology structure. Nodes represent OWL classes; edges are
object properties. The three intention-relevant relations
(\texttt{motionTowardEgo}, \texttt{spatialConfiguration}, \texttt{ttcRisk})
connect \texttt{SceneParticipant} to their respective enumeration classes.
\texttt{hasCoarseCategory} connects the underlying \texttt{Participant} to
\texttt{AgentCoarseType}.}
\label{fig:ontology-diagram}
\end{figure*}

Figure~\ref{fig:ontology-diagram} shows the complete class and property structure of the NIKG ontology; the subsections below describe each layer in turn.

\subsubsection{Baseline ontology}

NIKG inherits the class hierarchy \texttt{Dataset $\rightarrow$ Trip $\rightarrow$ Sequence $\rightarrow$ Scene} from nSKG \cite{mlodzianNuScenesKnowledgeGraph2023a}. It also inherits temporal links between scenes (\texttt{hasNextScene}, \texttt{hasPreviousScene}).

\texttt{Dataset} is the root individual for the exported \ac{kg} instance. It records the declared dataset name and version and links downward to every \texttt{Trip}.
Each \texttt{Trip} corresponds to one ego-centric recording unit packaged from nuScenes, aligned with one nuScenes scene, and owns exactly one \texttt{Sequence}.
A \texttt{Sequence} is the temporal backbone of that trip: the ordered chain of keyframes materialized as \texttt{Scene} nodes.
Each \texttt{Scene} is one discrete time slice. In the NIKG instantiation, it corresponds to one nuScenes sample chosen along the \texttt{CAM\_FRONT} keyframe chain, holds the observations at that instant, and connects forward and backward along the sequence via the temporal properties above.

Agent-level concepts separate \emph{persistent identity} from \emph{per-frame observation}, following the same pattern as nSKG \cite{mlodzianNuScenesKnowledgeGraph2023a}.
\texttt{Participant} denotes a physical agent with stable identity across scenes within the sequence (derived from nuScenes instance-level tracking).
\texttt{EgoVehicle} denotes the sensing ego vehicle itself; it is structurally distinct from traffic agents and is not represented as a traffic-facing \texttt{SceneParticipant}.
For every non-ego observer agent seen in a frame, the graph contains a \texttt{SceneParticipant}: one RDF individual whose role is to represent ``this participant \emph{as observed in this scene},'' linked to its \texttt{Participant} and enclosing \texttt{Scene} via \texttt{isSceneParticipantOf} and \texttt{hasSceneParticipant}, with centroid coordinates attached through \texttt{Point2D} (normalized image-plane position on \texttt{CAM\_FRONT}).

Scene participants are additionally typed using \emph{coarse} labels derived from nuScenes detection categories, rather than enumerating the full fine-grained taxonomy as closed \ac{owl} individuals.
Each non-ego \texttt{SceneParticipant} receives exactly one of four buckets---\emph{Vehicle}, \emph{Pedestrian}, \emph{MovableObject}, or \emph{Infrastructure}---via a fixed mapping from raw annotation category strings.
This keeps the \ac{tbox} compact (four coarse-type individuals instead of dozens of nuScenes leaf categories), simplifies schema maintenance and typical queries, and matches high-level intention analysis; the original nuScenes category string may still be emitted as literal metadata on the same node where available.

\subsubsection{Intention-relevant relations}
\label{subsec:intention-relations}

NIKG extends every \texttt{SceneParticipant} with three \emph{new} object properties that are not part of the inherited nSKG presentation \cite{mlodzianNuScenesKnowledgeGraph2023a}: \texttt{motionTowardEgo}, \texttt{spatialConfiguration}, and \texttt{ttcRisk}.
They attach exclusively to \texttt{SceneParticipant} nodes because each label is defined relative to ego pose at that keyframe.
Each property uses a closed enumerated range so that instance values are machine-checkable and query-friendly.
Table~\ref{tab:properties} summarises the properties, their ranges, and their legal values.

\begin{table}[t]
\centering
\footnotesize
\caption{Intention-relevant object properties introduced by NIKG.}
\label{tab:properties}
\begin{tabular}{@{}p{0.23\linewidth}p{0.19\linewidth}p{0.16\linewidth}p{0.32\linewidth}@{}}
\toprule
Property & Domain & Range & Values \\
\midrule
\texttt{motionTowardEgo} & \texttt{SceneParticipant} & \texttt{MotionState} & Approaching, Receding, Stationary \\
\texttt{spatialConfiguration} & \texttt{SceneParticipant} & \texttt{SpatialZone} & Preceding, Following, Left/Right $\times$ \{Preceding, Alongside, Following\} \\
\texttt{ttcRisk} & \texttt{SceneParticipant} & \texttt{TTCRiskLevel} & High, Medium, Low \\
\bottomrule
\end{tabular}
\end{table}

The paragraphs below specify operational semantics---inputs, thresholds, and emission rules---shared between the \ac{tbox} enumerations and the pipeline implementation (stage~4).

\textbf{Motion toward ego} (\texttt{motionTowardEgo}). For each participant
with a previous observation, the change in planar distance to the
ego vehicle between consecutive keyframes is computed. If the magnitude of the distance
change is below a fixed threshold, the participant is labeled
\texttt{Stationary}; otherwise it is labeled \texttt{Approaching} or
\texttt{Receding} depending on the sign. The threshold is fixed at 0.5~m.

\textbf{Spatial configuration} (\texttt{spatialConfiguration}). Each
participant is projected into the ego body frame and assigned to one of eight
surround zones, with thresholds of 3.0~m longitudinally, 1.75~m laterally,
and a 50~m maximum range. The resulting label provides a symbolic description
of local surround layout \cite{deoConvolutionalSocialPooling2018}.

\textbf{\Ac{ttc} risk} (\texttt{ttcRisk}). A one-step closing-speed approximation
yields a classical \ac{ttc} estimate at each keyframe ($\Delta t = 0.5$~s).
The same discrete risk buckets as \cite{manzourVehicleLaneChange2023} are adopted:
\texttt{High} when $\mathrm{TTC}<4$~s, \texttt{Medium} when $4~\mathrm{s}\leq \mathrm{TTC}<10$~s, and \texttt{Low} when $\mathrm{TTC}\geq 10$~s.
The relation is not emitted on first appearance or when closing speed is negligible \cite{haywardTimeCollisionTechnique1972}.

\subsubsection{Self-describing artifact}

At export time the pipeline merges the \ac{tbox} ---the class and property
declarations---with the \ac{abox} ---the generated instance triples---into a single
\ac{ttl} file under one namespace.
The resulting artifact is self-describing:
any consumer loading the file obtains both the vocabulary and the data without
a separate schema lookup, following the release pattern of nSKG \cite{mlodzianNuScenesKnowledgeGraph2023a}.

\subsection{Knowledge graph generation pipeline}
\label{sec:pipeline}

This subsection covers how those ontology semantics are instantiated on nuScenes: the stages from raw records to populated predicates (Section~\ref{subsec:intention-relations}), the minimal input contract, and reproducibility of the exported artifact.

\subsubsection{Overview and input contract}

\begin{figure}[t]
\centering
\begin{tikzpicture}[
    node distance = 3.0mm and 3.5mm,
    box/.style = {
        rectangle, rounded corners=2pt, draw=black!70, fill=gray!10,
        text width=2.25cm, minimum height=1.65cm,
        align=center, font=\scriptsize
    },
    arrow/.style = {-{Stealth[length=4pt]}, thick, draw=black!60}
]

\node[box] (ingest) {%
    \textbf{1. Ingest}\\[2pt]
    ego pose\\
    annotations\\
    instance ids\\
    keyframe order};

\node[box, right=of ingest] (records) {%
    \textbf{2. Records}\\[2pt]
    \texttt{KeyframeRecord}\\
    \texttt{SceneBundle}};

\node[box, right=of records] (domain) {%
    \textbf{3. Domain}\\[2pt]
    \texttt{Dataset}\\
    \texttt{Trip / Sequence}\\
    \texttt{Scene}\\
    \texttt{ScenePartic.}};

\node[box, right=of domain] (relations) {%
    \textbf{4. Relations}\\[2pt]
    \texttt{motionTowardEgo}\\
    \texttt{spatialConfig.}\\
    \texttt{ttcRisk}};

\node[box, right=of relations] (export) {%
    \textbf{5. RDF Export}\\[2pt]
    ontology merge\\
    \ac{tbox} + \ac{abox}\\
    \texttt{.ttl} artifact};

\draw[arrow] (ingest)    -- (records);
\draw[arrow] (records)   -- (domain);
\draw[arrow] (domain)    -- (relations);
\draw[arrow] (relations) -- (export);

\end{tikzpicture}
\caption{NIKG generation pipeline. Each stage is independently testable; the final artifact merges schema and instance triples into a single self-describing Turtle file.}
\label{fig:pipeline}
\end{figure}
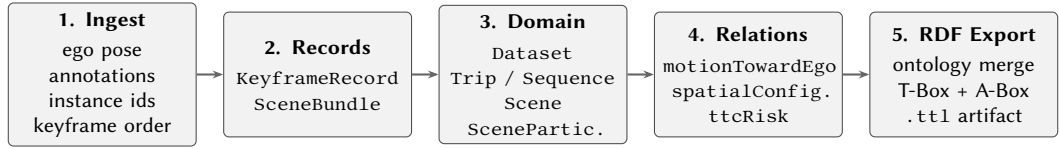

The generation pipeline has five stages, shown in Figure~\ref{fig:pipeline}.
Raw nuScenes records are fetched first (stage~1), then immediately converted
into lightweight neutral data structures that decouple the rest of the pipeline
from the nuScenes API (stage~2). These structures are assembled into the domain
object hierarchy---\texttt{Dataset}, \texttt{Trip}, \texttt{Sequence},
\texttt{Scene}, and \texttt{SceneParticipant}---without any relation labels
attached yet (stage~3). Relation computation then runs as a separate pass over
the assembled dataset, populating the three intention properties on each
\texttt{SceneParticipant} in place (stage~4). Finally, the exporter parses the
ontology and emits one RDF triple per populated field, merging schema and
instance data into the output \ac{ttl} file (stage~5).

The current implementation targets nuScenes and consumes only a minimal subset
of fields: ego pose in a shared planar frame, annotation positions, stable
instance identities, and keyframe ordering \cite{nuscenes}. Optional camera
intrinsics and images may be used for figures, but they are not required for
the core graph. Any ego-centric dataset exposing stable tracks, an ego pose,
and a shared planar reference frame could in principle be adapted with a thin
adapter. Table~\ref{tab:input-contract} summarises this minimal input contract.

\begin{table}[t]
\centering
\small
\caption{Minimal input contract for graph instantiation.}
\label{tab:input-contract}
\begin{tabular}{lll}
\toprule
Signal & Source & Used for \\
\midrule
Ego pose $(x,y,\psi)$ & ego record & ego frame + \ac{ttc} \\
Annotation position & annotation & participant geometry \\
Instance identifier & annotation & track continuity \\
Keyframe order & sample chain & temporal relations \\
Optional image/intrinsics & camera metadata & figures / future 2D \\
\bottomrule
\end{tabular}
\end{table}

The relation-population stage (pipeline stage~4) evaluates the three intention relations exactly according to Section~\ref{subsec:intention-relations}. It uses the signals listed in Table~\ref{tab:input-contract} to populate \texttt{motionTowardEgo}, \texttt{spatialConfiguration}, and \texttt{ttcRisk}.

\subsubsection{Export and reproducibility}

At export time, the pipeline first parses the ontology, then emits the instance
triples into the same graph under a single namespace, producing a single
self-describing \ac{ttl} file. Thresholds are fixed in code rather than exposed
as experiment flags, so the paper's numbers are reproducible from a single
command. The CLI accepts only the dataset root and ontology path:

\begin{verbatim}
python main.py --dataroot data/nuscenes \
               --version v1.0-mini \
               --ontology ontologies/ontology.ttl \
               --out-ttl graphs/multi_scene.ttl
\end{verbatim}

The full pipeline, ontology, and evaluation scripts are released as a public
companion repository~\cite{nikgRepo2026}. The
environment is frozen via \texttt{environment.yml} (conda, Python~3.11).

\section{Graph Validation}
\label{sec:validation}

The generated graph is examined along four axes: a concrete scene study that traces annotation labels through four consecutive keyframes, followed by quantitative checks on scale and coverage, semantic consistency, and temporal label coherence.

\subsection{Scene study: scene-1100, frames 19--22}
\label{subsec:scene-study}

\begin{figure*}[t]
\centering
\begin{subfigure}[t]{0.47\textwidth}
    \includegraphics[width=\textwidth]{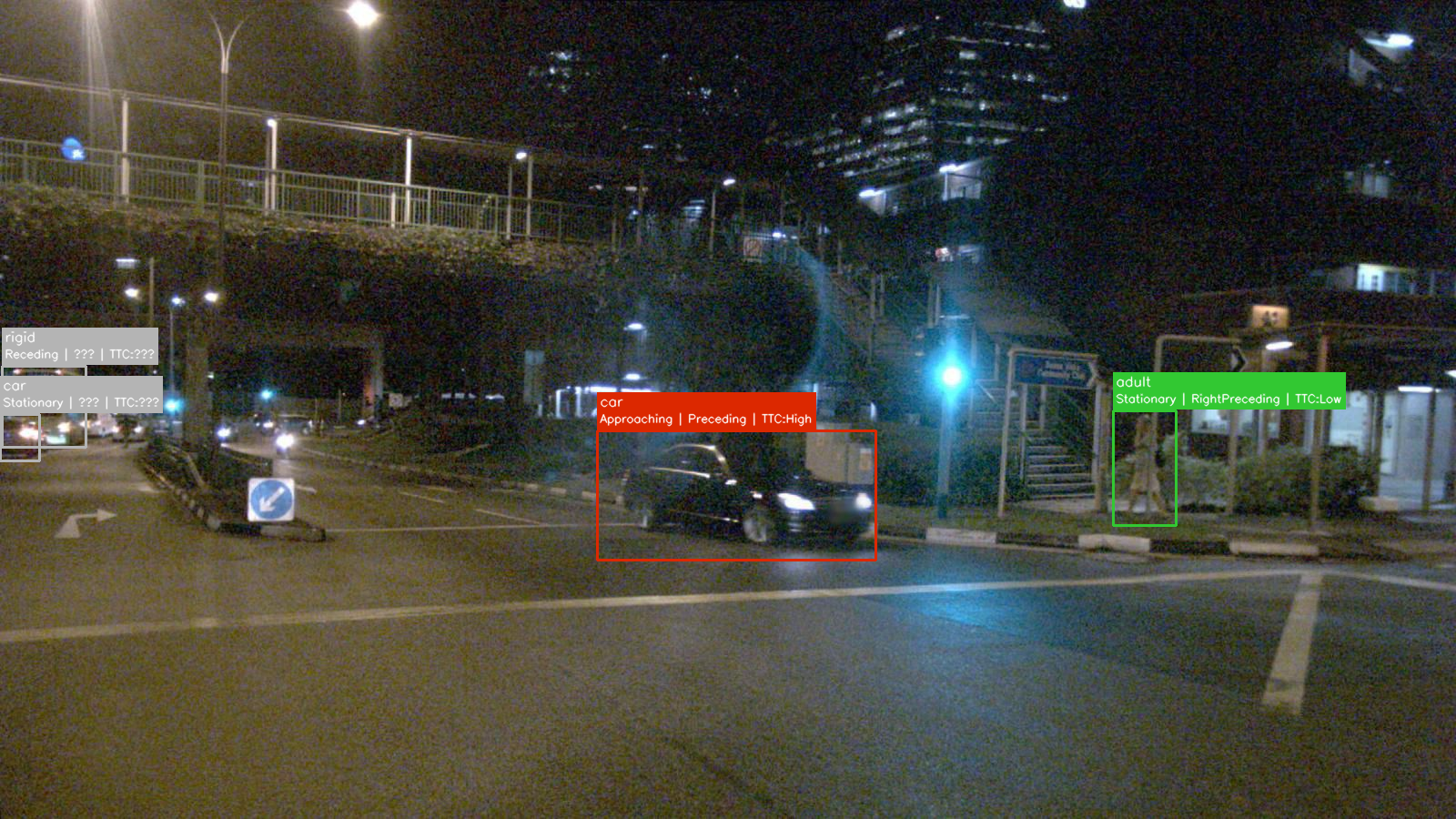}
    \caption{Frame~19: car 19.0~m ahead at \texttt{ttcRisk=High} (closing speed
    6.4~m/s, \ac{ttc}~2.97~s). Pedestrian at 8.4~m in \texttt{LeftFollowing}
    labelled \texttt{Stationary} (no TTC). Bus receding at 80.9~m beyond the
    50~m range cutoff: \texttt{Receding} only, no spatial or TTC label.}
    \label{fig:frame19}
\end{subfigure}\hfill
\begin{subfigure}[t]{0.47\textwidth}
    \includegraphics[width=\textwidth]{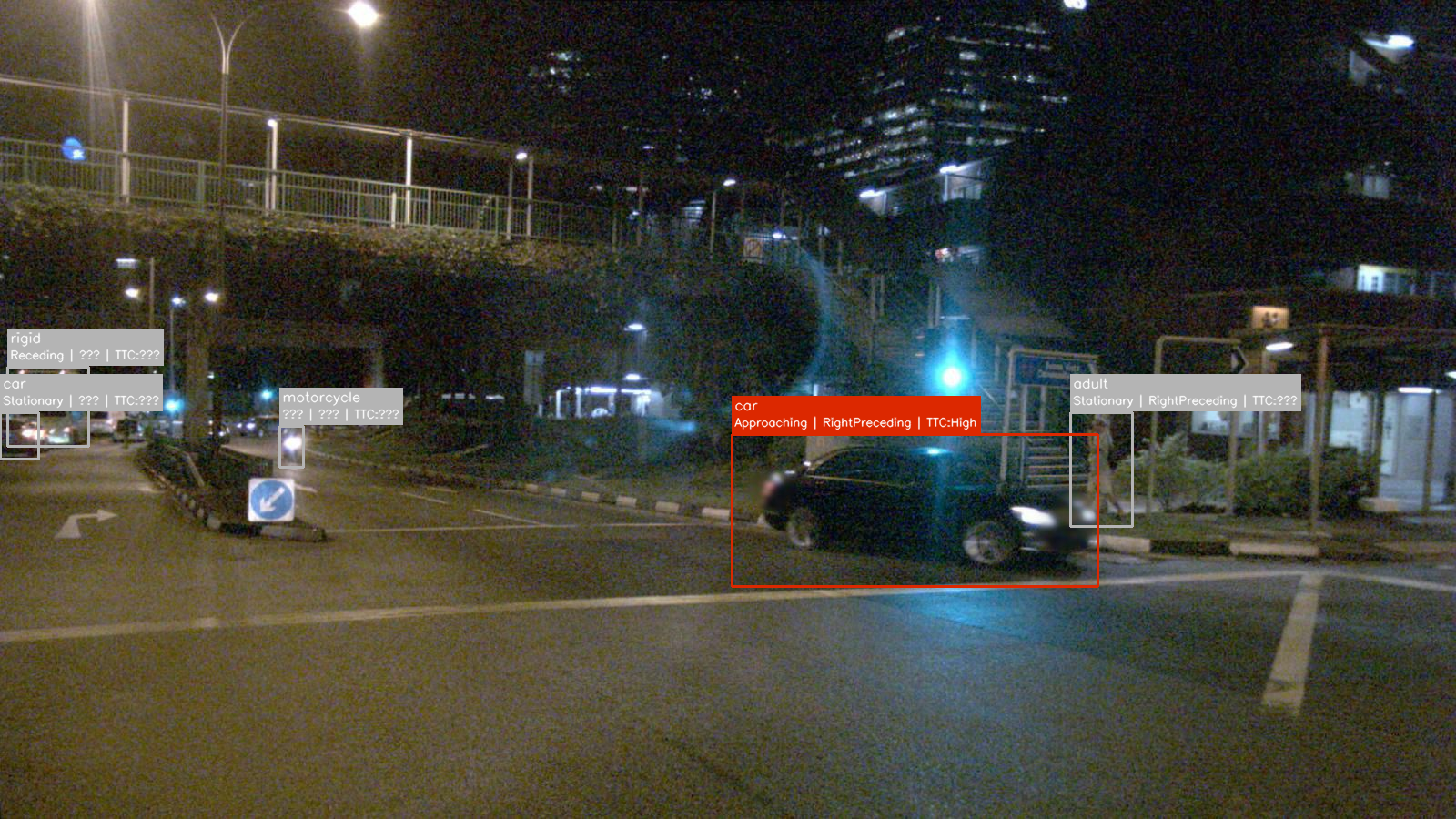}
    \caption{Frame~20: car closes to 16.9~m, \texttt{ttcRisk} remains
    \texttt{High}. Pedestrian and bus labels unchanged. A motorcycle enters
    the scene for the first time at 55.5~m (first appearance, beyond range
    cutoff): all three relation slots are empty.}
    \label{fig:frame20}
\end{subfigure}

\medskip

\begin{subfigure}[t]{0.47\textwidth}
    \includegraphics[width=\textwidth]{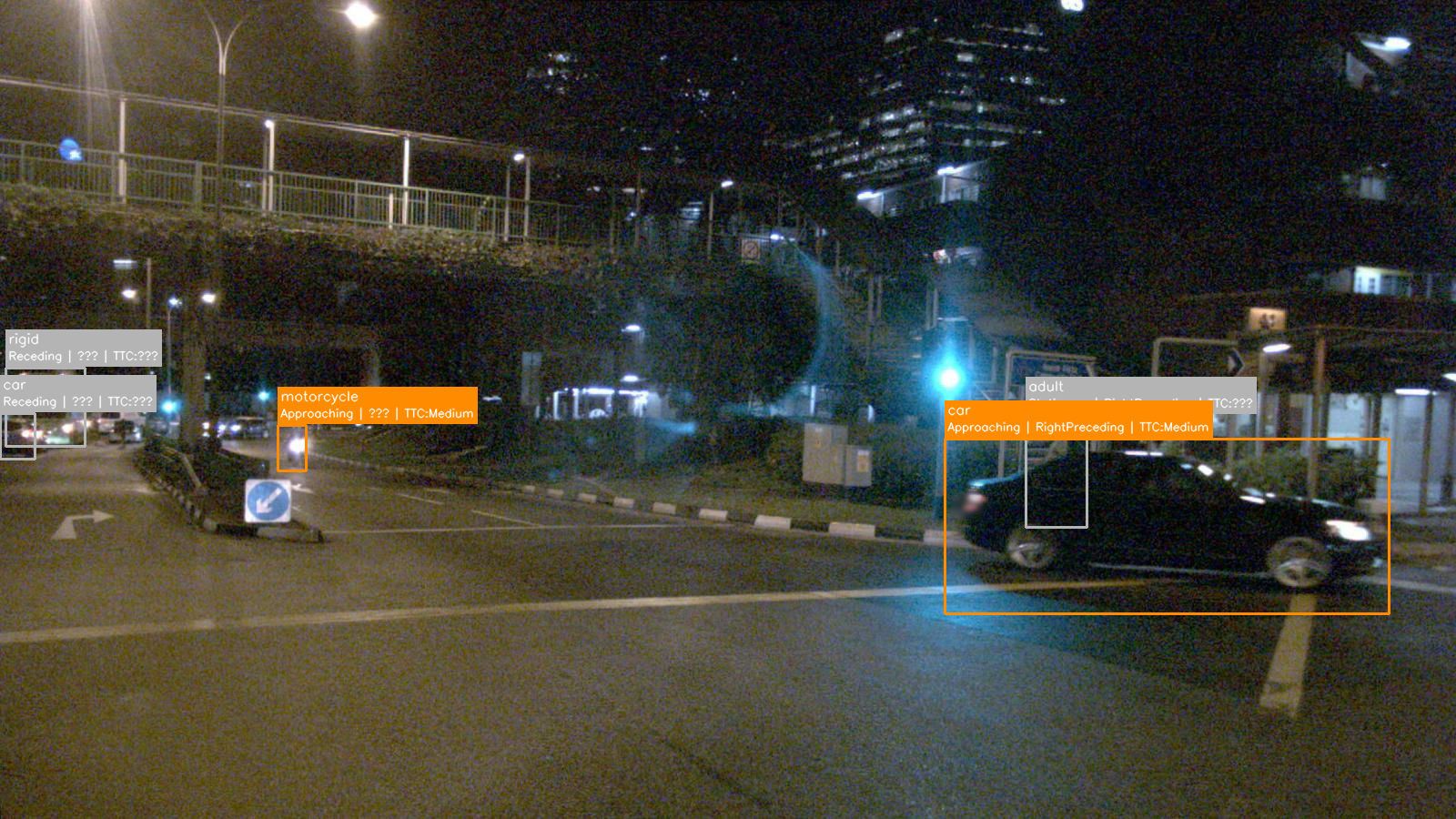}
    \caption{Frame~21: car continues closing, now at 15.6~m; deceleration brings \texttt{ttcRisk} down to
    \texttt{Medium} (\ac{ttc}~6.27~s). Pedestrian remains \texttt{Stationary};
    bus continues receding at 84.4~m. Motorcycle closes to 50.4~m and receives
    \texttt{ttcRisk=Medium}, but is still marginally outside the 50~m spatial
    range so \texttt{spatialConfiguration} is absent.}
    \label{fig:frame21}
\end{subfigure}\hfill
\begin{subfigure}[t]{0.47\textwidth}
    \includegraphics[width=\textwidth]{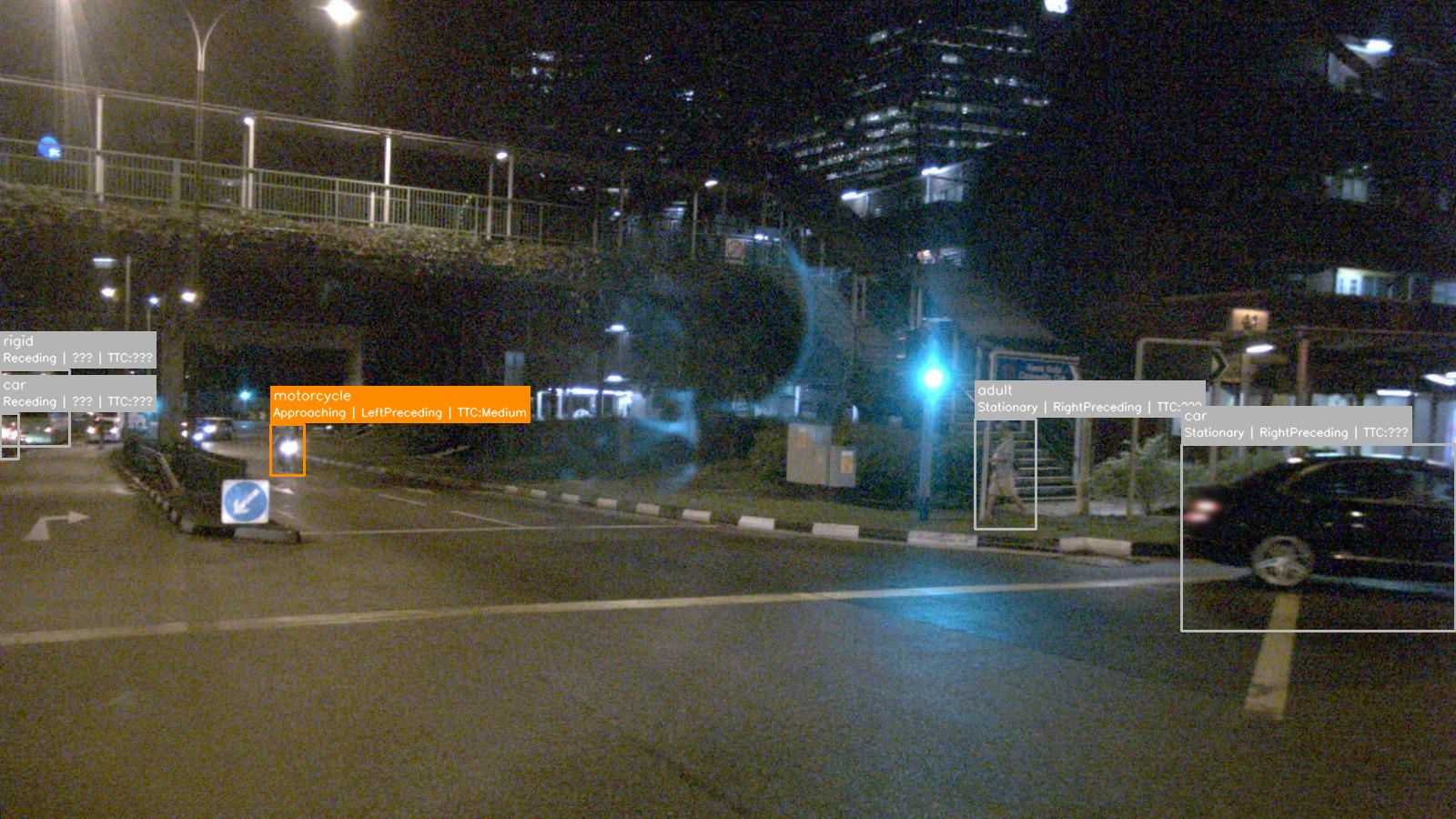}
    \caption{Frame~22: car brakes to near-stationary ($\Delta d = -0.046$~m),
    \texttt{ttcRisk} suppressed. Pedestrian and bus labels unchanged across all
    four frames. Motorcycle crosses into the 50~m zone at 45.9~m and is fully
    annotated: \texttt{Approaching}~| \texttt{LeftPreceding}~| \texttt{Medium}.}
    \label{fig:frame22}
\end{subfigure}
\caption{Scene-1100, frames~19--22 (0.5~s apart). Bounding boxes are colour-coded
by \texttt{ttcRisk}: red = \texttt{High}, orange = \texttt{Medium}, green = \texttt{Low},
grey = none. Each box label: coarse category / \texttt{motionTowardEgo}~|
\texttt{spatialConfiguration}~| \texttt{ttcRisk}.}
\label{fig:scene-study-frames}
\end{figure*}

Four consecutive keyframes (frames~19--22, 1.5~s at the 2~Hz keyframe rate)
from scene-1100 of nuScenes mini are examined to ground the annotation scheme
in a concrete traffic scenario.
Figure~\ref{fig:scene-study-frames} shows the annotated \texttt{CAM\_FRONT}
images; bounding boxes are colour-coded by \texttt{ttcRisk}
(red~=~\texttt{High}, orange~=~\texttt{Medium}, green~=~\texttt{Low},
grey~=~none) and each box carries a two-line overlay: coarse category,
then \texttt{motionTowardEgo}~|~\texttt{spatialConfiguration}~|~\texttt{ttcRisk}.

\textbf{Approaching car.}
A car 19.0~m ahead (\texttt{Preceding}) decelerates over the window.
Closing speed falls from 6.4~m/s in frame~19 (\ac{ttc}~2.97~s, \texttt{High})
to 4.3~m/s in frame~20 (\ac{ttc}~3.88~s, still \texttt{High}),
2.5~m/s in frame~21 (\ac{ttc}~6.27~s, \texttt{Medium}), and 0.09~m/s in
frame~22---just below the 0.1~m/s emission threshold---so no \texttt{ttcRisk}
triple is written and \texttt{motionTowardEgo} transitions to \texttt{Stationary}
despite the vehicle remaining at 15.6~m.
The arc (\texttt{High}, \texttt{High}, \texttt{Medium}, none) encodes the full
braking event as a sequence of instantaneous risk estimates.

\textbf{Stationary pedestrian.}
A pedestrian at 8.4~m in \texttt{LeftFollowing} carries \texttt{Stationary}
throughout; inter-frame deltas ($+$0.095, $+$0.041, $-$0.008, $-$0.007~m)
all lie within the 0.5~m band, and closing speeds (${\leq}$0.19~m/s) remain
below the TTC gate, so no \texttt{ttcRisk} triple is emitted.
This illustrates how the stationarity threshold absorbs localisation noise
without masking genuine displacement.

\textbf{Newly arriving motorcycle.}
The motorcycle is absent in frame~19, then enters the scene at 55.5~m in
frame~20 (first appearance, beyond the 50~m range cutoff): all three
relation slots are empty.
In frame~21 it has closed to 50.4~m---still marginally outside the range
boundary, so \texttt{spatialConfiguration} remains absent---but a closing
speed of 10.2~m/s yields \texttt{ttcRisk=Medium} (\ac{ttc}~4.95~s).
In frame~22 the participant crosses into the 50~m zone at 45.9~m and all
three relations are populated: \texttt{Approaching}, \texttt{LeftPreceding},
\texttt{ttcRisk=Medium} (\ac{ttc}~5.01~s).
The three-frame transition illustrates the sequenced effect of the
first-appearance rule, the range cutoff, and the closing-speed gate.

\textbf{Receding bus.}
A bus is tracked at 80.9--86.1~m across all four frames ($\Delta d \approx +1.5$--$+2.0$~m per step)
and consistently labelled \texttt{Receding}; the negative closing speed
suppresses \texttt{ttcRisk}, and the range beyond 50~m suppresses
\texttt{spatialConfiguration}.
Its uninterrupted presence in the graph confirms that all participants are
tracked and annotated regardless of range or approach direction; only the
relations whose emission conditions are satisfied are written.

\begin{figure}[t]
\centering
\includegraphics[width=0.92\linewidth]{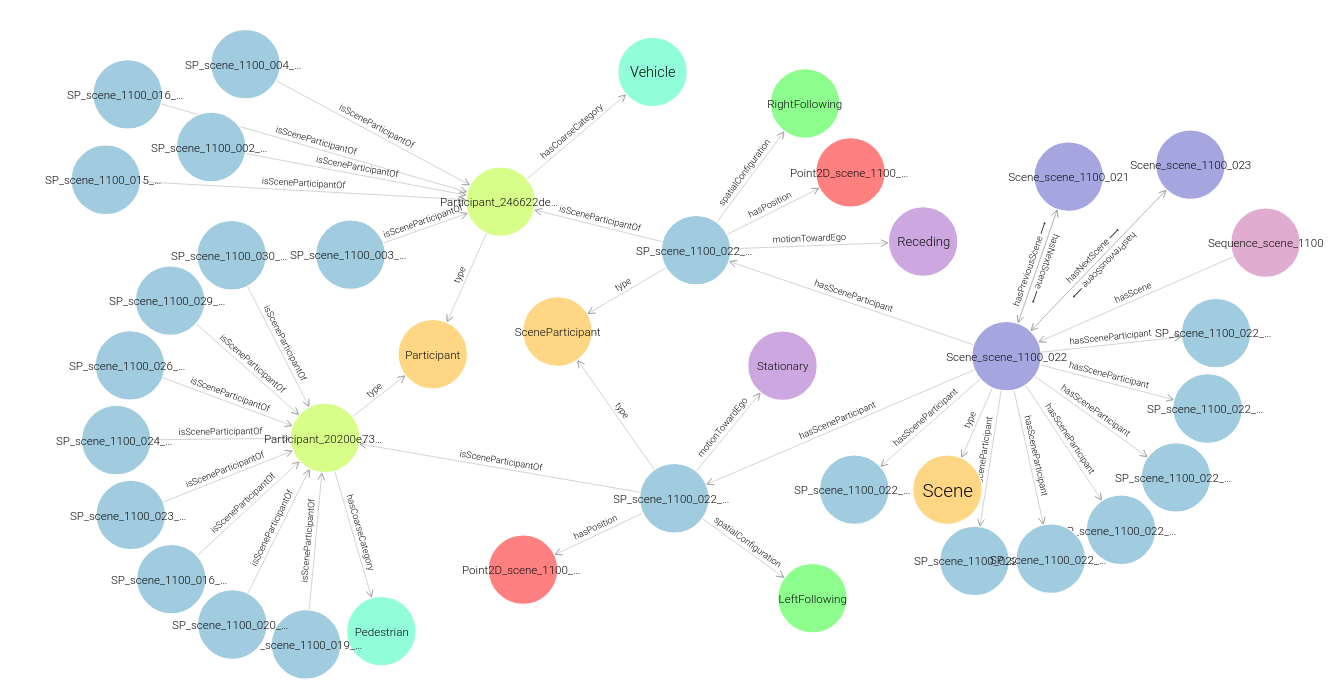}
\caption{Representative NIKG subgraph for scene-1100 frame~22. Two
\texttt{SceneParticipant} nodes are expanded to show their populated
intention-relation triples; the remaining 21 participants in the frame
are collapsed for clarity. The temporal chain linking this frame to
frames~21 and~23 is visible at the top of the diagram.}
\label{fig:scene-study-kg}
\end{figure}

Figure~\ref{fig:scene-study-kg} shows a representative subgraph for
frame~22; two \texttt{SceneParticipant} nodes are expanded to show their
intention-relation triples, while the remaining 21 participants are collapsed
for clarity.

\subsection{Scale and coverage}

Table~\ref{tab:coverage} summarises graph scale on nuScenes mini (18,538 \texttt{SceneParticipant} instances across 404 keyframes).
The artifact merges the NIKG \ac{tbox} with instance data and asserts
participant typing (\texttt{hasCoarseCategory}, \texttt{hasNuScenesCategory})
on each \texttt{SceneParticipant} in addition to the three intention-relevant
relations.
All eight \texttt{SpatialZone} values and all
three \texttt{TTCRiskLevel} values are instantiated, confirming that the
graph is non-degenerate and the full enumerated vocabulary is populated.

\begin{table}[t]
\centering
\small
\caption{Graph scale on nuScenes mini (10 scenes).}
\label{tab:coverage}
\begin{tabular}{lr}
\toprule
Statistic & Value \\
\midrule
Total triples                          & 211,335 \\
Scenes (keyframes)                     & 404 \\
Unique participants                    & 911 \\
\texttt{SceneParticipant} instances    & 18,538 \\
\midrule
\texttt{hasCoarseCategory} triples      & 18,538 \\
\texttt{hasNuScenesCategory} triples  & 18,538 \\
\texttt{motionTowardEgo} triples       & 17,627 \\
\texttt{spatialConfiguration} triples  & 15,166 \\
\texttt{ttcRisk} triples               & 8,156 \\
\bottomrule
\end{tabular}
\end{table}

The relation counts are each smaller than the total instance count for
defined reasons. \texttt{motionTowardEgo} requires a previous keyframe, so
first appearances carry no label; the difference of 911 equals exactly the
number of unique participants. \texttt{spatialConfiguration} excludes
agents beyond 50~m planar distance (3,372 out-of-range instances).
\texttt{ttcRisk} is only emitted when closing speed exceeds 0.1~m/s;
receding and stationary agents account for approximately 10,400 of the
unlabelled instances.

\subsection{Semantic consistency}

Table~\ref{tab:consistency} reports three implications that test coherence among the intention relations: each row states a premise and a conclusion expected whenever that premise appears on a \texttt{SceneParticipant} keyframe in the RDF graph. Operationally, triples are scanned and instances where the premise holds are collected; \emph{Total} is that premise count, \emph{Pass} is how many also satisfy the conclusion, and \emph{Rate} is $\mathrm{Pass}/\mathrm{Total}$. The rate therefore answers ``among agents that trigger this premise, what fraction also satisfy the consequent?''---near 100\% means the labels agree under that rule; a substantially lower rate flags systematic tension between features (often from geometric modelling rather than a malformed export).

The three rules follow directly from the operational definitions in Section~\ref{subsec:intention-relations}.
\emph{(i)}~\texttt{ttcRisk=High} and \texttt{motionTowardEgo} both encode closing behavior via different surrogates (one-step \ac{ttc} bin vs.\ signed distance delta), so \texttt{High} should align with \texttt{Approaching} rather than \texttt{Stationary} or \texttt{Receding}.
\emph{(ii)}~Because \texttt{ttcRisk} uses planar line-of-sight distance, imminent closure should generally correspond to an ahead-of-ego surround cell (\texttt{Preceding}, \texttt{LeftPreceding}, or \texttt{RightPreceding}); mismatches surface the gap between LOS and path-aware estimates.
\emph{(iii)}~\texttt{Stationary} (inter-frame $|\Delta d| < 0.5$~m) and \texttt{ttcRisk=High} (strong closing-speed signal) are contradictory by construction: co-occurrence indicates threshold-band tension between two independently computed labels.

\begin{table}[t]
\centering
\small
\caption{Semantic consistency checks. Pass rate = fraction of \emph{premise}
instances that also satisfy the conclusion.}
\vspace{2pt}
\label{tab:consistency}
\begin{tabular}{lrrr}
\toprule
Constraint & Total & Pass & Rate \\
\midrule
\texttt{ttcRisk=High} $\Rightarrow$ \texttt{Approaching}
    & 1{,}990 & 1{,}988 & 99.9\% \\
\texttt{ttcRisk=High} $\Rightarrow$ \texttt{spatialConfiguration} in forward zone
    & 1{,}892 & 1{,}688 & 89.2\% \\
\texttt{Stationary} $\Rightarrow$ \texttt{ttcRisk} $\neq$ \texttt{High}
    & 3{,}784 & 3{,}782 & 99.9\% \\
\bottomrule
\end{tabular}
\end{table}

Rows~\emph{(i)} and~\emph{(iii)} are near-perfect (both 99.9\%). The two violations in
each are the same pair of agents: both carry \texttt{Stationary} and
\texttt{RightAlongside} labels with \texttt{ttcRisk=High}. In both cases
the distance delta between frames falls within the 0.5~m \texttt{Stationary}
band, but the absolute distance at the current frame is small enough to
produce a sub-4~s \ac{ttc} estimate (hence \texttt{ttcRisk=High} under the $<$4~s bin). This is a threshold-band collision between
two independently computed labels rather than a pipeline error.

Row~\emph{(ii)} passes at 89.2\%. Forward zones (\texttt{Preceding},
\texttt{LeftPreceding}, \texttt{RightPreceding}) are expected for most
high-risk agents, but 204 instances are in lateral or rear zones. This
occurs because \texttt{ttcRisk} uses line-of-sight distance rather than a
path-aware estimate: a vehicle closing from the side or rear can produce a
decreasing planar distance and thus a high \ac{ttc} estimate. This is consistent
with the limitation noted in Section~\ref{sec:limits}. The 98 high-risk
agents without a \texttt{spatialConfiguration} label (beyond the 50~m range
cut) are excluded from the denominator.

\subsection{Temporal label coherence}

For each \texttt{SceneParticipant} with a \texttt{motionTowardEgo} label at
scene $t$, the observed distance change between $t$ and $t{+}1$ is tested
for consistency with that label: \texttt{Approaching} predicts a
decrease, \texttt{Receding} predicts an increase, and \texttt{Stationary}
predicts $|\Delta d| < 0.5$~m. Because the check uses the step \emph{after}
the one used to generate the label, it is non-circular.

Overall coherence is 92.5\% across 16,746 labeled transitions.
Per-label rates are: \texttt{Receding} 99.1\%, \texttt{Approaching} 91.0\%,
and \texttt{Stationary} 84.7\%. \texttt{Stationary} is lowest because it
occupies the narrowest distance band; agents near the 0.5~m boundary at $t$
frequently cross it at $t{+}1$. Errors are almost entirely adjacent-label
transitions (\texttt{Approaching}$\leftrightarrow$\texttt{Stationary},
\texttt{Stationary}$\leftrightarrow$\texttt{Receding}); cross-direction
errors (\texttt{Approaching}$\to$\texttt{Receding} or vice versa) account
for only 158 of 16,746 transitions (0.9\%).

\section{Conclusion}

This paper presents NIKG: a self-describing \ac{owl} ontology, a reproducible nuScenes pipeline, and a graph-characterization analysis that makes ego-relative motion, surround configuration, and \ac{ttc} risk explicit as first-class relations on each observed agent.
Validation on 18,538 \texttt{SceneParticipant} instances shows 92.5\% forward motion coherence and near-perfect pass rates on two of three semantic consistency implications; the 89.2\% rate on the forward-zone check reflects a known design trade-off (line-of-sight \ac{ttc} vs path-aware closure) that the current schema handles by design rather than obscuring.
By formalizing the contextual factors that the maneuver-intention literature consistently relies on, NIKG provides a reusable semantic layer that can support downstream intention-prediction work through \ac{sparql}-based feature extraction, \acp{kge}, or hybrid neuro-symbolic pipelines.

\section{Limitations and Future Work}
\label{sec:limits}

All three intention-relevant properties are defined relative to the ego vehicle; pairwise agent-to-agent relations (e.g.\ following, yielding) fall outside the current schema and would require extending the ontology with agent-pair subjects.
Portability to other ego-centric datasets has not been demonstrated empirically: the graph is instantiated and evaluated on nuScenes mini only.
The primary direction for future work is connecting NIKG to a downstream intention-prediction task via \ac{sparql}-based feature extraction, \ac{kge}, or a hybrid neuro-symbolic pipeline.

\section*{Declaration on Generative AI}
During the preparation of this work, the author(s) used Cursor (agent mode) and
Prism in order to: grammar and spelling check; paraphrase and reword; improve
writing style; peer review simulation; formatting assistance
(Tables~\ref{tab:properties}, \ref{tab:input-contract}, \ref{tab:coverage},
and~\ref{tab:consistency}); and citation management.
Further, the author(s) used Cursor (agent mode) for Figures~\ref{fig:pipeline-position}
and~\ref{fig:pipeline} in order to: generate images.
After using these tool(s)/service(s), the author(s) reviewed and edited the
content as needed and take(s) full responsibility for the publication's content.

\bibliography{prism-uploads/Masters}

\end{document}